%% file: bare_jrnl.tex
\documentclass[journal]{IEEEtran}
\IEEEoverridecommandlockouts
\usepackage{graphicx}
\def\BibTeX{{\rm B\kern-.05em{\sc i\kern-.025em b}\kern-.08em
    T\kern-.1667em\lower.7ex\hbox{E}\kern-.125emX}}
\usepackage[font=small,labelfont=bf,tableposition=top]{caption}

\usepackage{blindtext}
\usepackage{caption}

\let\oldtwocolumn\twocolumn
\renewcommand\twocolumn[1][]{%
    \oldtwocolumn[{#1}{
    \begin{center}
    \vskip-3ex
        \centering
        \includegraphics[width=0.99\textwidth]{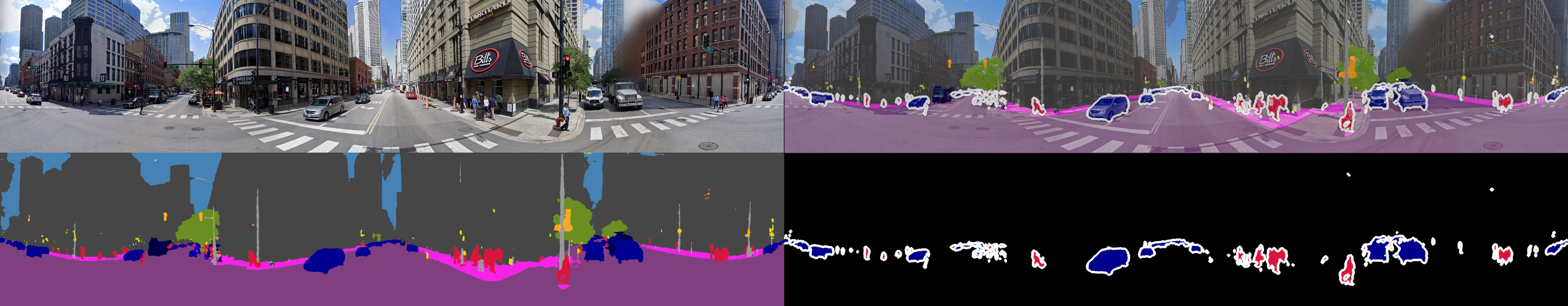}
        \captionof{figure}{Within this work, we differentiate between various levels of image understanding: The original image (first row, left) can be interpreted as a panoramic semantic map (second row, left) by assigning a label to each pixel without differentiating between different instances of countable objects. Instances of countable objects are distinguished in the panoramic instance understanding (second row, right). The panoramic panoptic understanding (first row, right), which is the proposed method in this paper, builds on top of the previous understandings by eliminating their shortcomings: If possible different instances are distinguished and we guarantee that a label is assigned to each pixel.}
        \label{fig.introduce}
    \end{center}
    }]
}

\usepackage{amsmath,amssymb,amsfonts}
\usepackage[inline]{enumitem}
\usepackage{algorithmic}
\usepackage{graphicx}
\usepackage{textcomp}

\usepackage{booktabs}
\usepackage{multirow}
\usepackage[breaklinks,colorlinks]{hyperref}
\hypersetup{colorlinks=true,linkcolor=red,citecolor=blue,urlcolor=magenta}

\usepackage{pgfplots} %Added by Alex

\usepackage[utf8]{inputenc}
\usepackage[T1]{fontenc}
\usepackage{float}
\usepackage{adjustbox}
\usepackage{array}
\usepackage{etoolbox}
\usepackage{dsfont}
\usepackage{xcolor}
\definecolor{revised_color}{HTML}{0066CC}

\usepackage{subcaption}

\usepackage[capitalise]{cleveref} 
\usepackage{tikz}
\usetikzlibrary{datavisualization}
\begin{document}
\title{Panoramic Panoptic Segmentation: Insights Into Surrounding Parsing for Mobile Agents via Unsupervised Contrastive Learning}
%\title{Panoramic Panoptic Segmentation: Insights Into Mobile Surrounding Understanding via Unsupervised Contrastive Learning}

%\title{Panoramic panoptic segmentation: Towards complete surrounding understanding via unsupervised contrastive learning}

\author{Alexander Jaus$^{1}$, Kailun Yang$^{1}$, and Rainer Stiefelhagen$^{1}$
\thanks{This work was supported in part by the Federal Ministry of Labor and Social Affairs (BMAS) through the AccessibleMaps project under Grant 01KM151112, in part by the University of Excellence through the ``KIT Future Fields'' project, in part by the Competence Center Karlsruhe for AI Systems Engineering (CC-KING, \url{www.ai-engineering.eu}) sponsored by the Ministry of Economic Affairs, Labour and Housing Baden-W{\"u}rttemberg, in part by the Helmholtz Association Initiative and Networking Fund on the HAICORE@KIT partition, and in part by Hangzhou SurImage Technology Company Ltd.
\textit{(Corresponding author: Kailun Yang.)}}
\thanks{$^{1}$Authors are with Institute for Anthropomatics and Robotics, Karlsruhe Institute of Technology, Germany (e-mail: alexander.jaus@kit.edu, kailun.yang@kit.edu, rainer.stiefelhagen@kit.edu).}
\thanks{The WildPPS dataset, code implementations, and model weights will be made publicly available at \url{https://github.com/alexanderjaus/PPS}}
}


\markboth{IEEE Transactions on Intelligent Transportation Systems, June~2022}%
{Jaus \MakeLowercase{\textit{et al.}}: Panoramic Panoptic Segmentation}



\maketitle

\begin{abstract}
\input{Tex_content/Abstract}
\end{abstract}


% \fi
\IEEEpeerreviewmaketitle

\section{Introduction}
\input{Tex_content/Introduction}

\section{Related Work}
\input{Tex_content/Related_work}

\section{Proposed Framework}
\input{Tex_content/Proposed_Framework}

\section{Experiments and Discussions}
\input{Tex_content/Experiments}

\section{Conclusion}
\input{Tex_content/Conclusion}

\ifCLASSOPTIONcaptionsoff
  \newpage
\fi




\bibliographystyle{IEEEtran}

\bibliography{Bibliography}
\end{document}

%% file: Tex_content/Abstract.tex
In this work, we introduce panoramic panoptic segmentation, as the most holistic scene understanding, both in terms of Field of View (FoV) and image-level understanding for standard camera-based input. A complete surrounding understanding provides a maximum of information to a mobile agent. This is essential information for any intelligent vehicle to make informed decisions in a safety-critical dynamic environment such as real-world traffic. In order to overcome the lack of annotated panoramic images, we propose a framework which allows model training on standard pinhole images and transfers the learned features to the panoramic domain in a cost-minimizing way. The domain shift from pinhole to panoramic images is non-trivial as large objects and surfaces are heavily distorted close to the image border regions and look different across the two domains. Using our proposed method with dense contrastive learning, we manage to achieve significant improvements over a non-adapted approach. Depending on the efficient panoptic segmentation architecture, we can improve $3.5{-}6.5\%$ measured in Panoptic Quality (PQ) over non-adapted models on our established Wild Panoramic Panoptic Segmentation (WildPPS) dataset. Furthermore, our efficient framework does not need access to the images of the target domain, making it a feasible domain generalization approach suitable for a limited hardware setting. As additional contributions, we publish WildPPS: The first panoramic panoptic image dataset to foster progress in surrounding perception and explore a novel training procedure combining supervised and contrastive training.

%% file: Tex_content/Introduction.tex
\label{Ch: Introduction}
\IEEEPARstart{P}{anoptic} segmentation is the so far most complete segmentation task to describe the context of an image~\cite{kirillov2019panoptic}.
It seamlessly addresses stuff and thing classes and thus unifies semantic segmentation which does not differentiate between instances and instance segmentation which falls short to segment uncountable objects. 
Both are critical pieces of information for the task of scene understanding in an autonomous driving setting. Not distinguishing between different instances of cars or pedestrians does not allow to anticipate the dynamics of individuals as they tend to interact with each other. 

Typically, in a street-scene context, stuff classes such as \emph{roads} or \emph{sidewalks} can be used to find traversable areas, whereas \emph{cars} or \emph{pedestrians} which are represented by the thing class, can be interacted with by the mobile agent in order to protect vulnerable road users or avoid moving obstacles. A purely semantic segmentation approach~\cite{yang2019pass,zhang2022bending}, which fails to identify the number of instances of the thing class, is insufficient, as the number of pedestrians or cars is an important piece of information in order to plan in accordance with the situation~\cite{peng2020deep}. 

Despite the high level of information provided by panoptic image understanding on standard pinhole images, as it is the de facto center of research at the moment, these works lack a crucial source of information which comes from the limited Field of View (FoV) of the input image. As real-world traffic is a highly dynamic environment in which a lot of movement is happening around the participants, the problem of people or vehicles moving out of the FoV of the mobile agent is easily imaginable~\cite{yang2019pass}. This poses severe problems due to the lack of information containing the entire surrounding and the inability of the agent to make proper decisions which may even lead to accidents~\cite{zhang2021exploring}. Thus, both pieces of information are equally important: the image should cover the entire surrounding and should be understood holistically~\cite{jaus2021panoramic}. 

\begin{figure}
\centering
\begin{tikzpicture}
\begin{axis}[
    xlabel={Field of View},
    ylabel={Panoptic Quality (PQ)},
    xmin=140, xmax=340,
    ymin=52, ymax=64,
    xtick={160,240,320},
    xticklabel=$\pgfmathprintnumber{\tick}^\circ$,
    ytick={52, 55, 58, 61, 64},
    ymajorgrids=true,
    grid style=dashed,
    width=0.3\textwidth,
    legend style={at={(0.835,1)},anchor=north east}
]

\addplot[
    color=black,
    mark=square,
    ]
    coordinates {
    (140,55.55)(170,56.14)(205,57.10)(237,57.26)(271,58.65)(304,59.73)(338,63.27)
    };
    \addlegendentry{Robust Model}

\addplot[
    color=black,
    mark=triangle,
    ]
    coordinates {
    (140,54.03)(170,53.19)(205,54.13)(237,53.08)(271,54.98)(304,56.11)(338,60.25)
    };
    \addlegendentry{Baseline Model}

\end{axis}
\end{tikzpicture}
%\vskip-1ex
\caption{Comparison of a standard Seamless Scene Segmentation model~\cite{porzi2019seamless} to its robust counterpart depending on the Field of View (FoV) on the WildPPS dataset. Larger FoVs tend to widen the gap between the two models confirming the effectiveness of the proposed Panoramic Robust Feature (PRF) framework.}
\label{fig: Standard_robust_comparison}
\vskip-3ex
\end{figure}
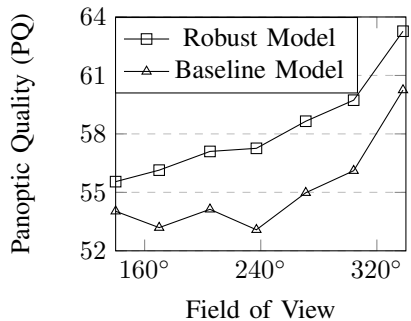

%Problems of other solutions
One approach to overcome the aforementioned FoV problem is to use multi-camera or LiDAR systems~\cite{berrio2020camera,peng2022mass,testolina2022selma}. These systems, however, introduce a further level of complexity as they either need to stitch images of multiple camera sources or resort to the combination of the RGB image information with LiDAR views in order to get a holistic view of the entire surrounding~\cite{petrovai2022semantic_camera}.
Within this work, we aim to introduce a simple, efficient, and comprehensive approach, which is suitable for a mobile agent setup.

%Introduce our approach
To this intent, we introduce Panoramic Panoptic Segmentation, to enable complete image understanding based on standard camera input.
There are a multitude of tasks which need to be addressed in order to properly operate mobile agents within a real-world scenario such as localization, motion planning, or robot controls. The one problem which these tasks have in common is that they rely on an interpretation of sensor data in order to collect information about the mobile agent as well as the world~\cite{ye2022pvo,gosala2022bird}. While there are multiple approaches to receive this information, this work focuses on scene understanding via standard camera input to derive information about the current state.
Our task combines panoptic image level understanding and the wide FoV of panoramic images~\cite{gao2022review}, tackling both previously discussed problems in a cost minimizing way because besides pure performance, a major design requirement for the derived approach is hardware and energy efficiency.

It is well known that segmentation models require a substantial amount of annotated data in order to be trained in a supervised fashion~\cite{cordts2016cityscapes}. However, dense, pixel-wise annotations are notoriously time-consuming and extremely expensive to produce, especially for panoramic images with a large FoV and many small objects co-occurring in the complex surroundings.

Due to this lack of available annotated panoramic images, we are forced to train our segmentation models on standard pinhole images, which introduces a domain shift to the proposed panoramic panoptic image segmentation task. A common problem when training and testing under different image distributions is the performance drop of models in the unseen target domain~\cite{yang2019pass}. 
The domain shift from pinhole- to panoramic images is no exception. The differences between standard pinhole images and panoramic images, as shown in Fig.~\ref{fig.introduce}, are quite apparent. Streets span through the entire image, whereas sidewalks are no longer present only on the side and now form bow-shaped islands in central parts of the image. These properties have not been observed by the model during the training and make their correct segmentation challenging. As they represent the traversable areas within a street scene, it is crucial to segment them in a correct way in order to avoid accidents or navigation failures.
In order to overcome these difficulties, we propose the Panoramic Robust Feature (PRF) framework which allows us to generate robust backbones via a contrastive pretext task. Contrastive learning does not only encourage similar features to be represented in a similar manner but more important, it pushes dissimilar features away from each other~\cite{he2020momentum,wang2020dense}. This leads to well separated clusters in the latent space of the backbone which proves to mitigate distribution shift performance drops.
The robust backbones can be inserted into the target panoptic model and the model can be trained in a standard supervised fashion on pinhole images. We find that our proposed approach significantly outperforms non-adapted models in a variety of segmentation scenarios. 

It furthermore comes with the advantage that we can make use of any large-scale dataset, as our method works best in our experiments when using the same domain for pretraining and supervised training. This does not only eliminate the need for labels within the target domain during training but even for images in the target domain making our approach feasible for domain generalization problems. Diverse autonomous transportation application scenarios, in which it may even be difficult to obtain more images from the target domain due to the need for specialized equipment, could benefit from the proposed method as well.
By adapting target models via our proposed PRF framework, we achieve significant improvements of over $6\%$ overall and over $7\%$ for the difficult stuff class measured in Panoptic Quality (PQ) over non-adapted baseline models on our extended panoramic panoptic dataset WildPPS~\cite{jaus2021panoramic}. In particular, the aforementioned challenging classes \emph{road} and \emph{sidewalk} on which non-adapted models perform poorly are now well addressed, which can be seen in Fig.~\ref{fig:Qualitative results}.
PRF consistently brings generalization gains on different FoVs cropped from the panoramas, as illustrated in Fig.~\ref{fig: Standard_robust_comparison}.
The proposed approach, thereby, attains state-of-the-art performance on both the public PASS~\cite{yang2019pass} and our established WildPPS benchmarks.

Finally, we discuss the feasibility of our efficient implementations of various state-of-the-art panoptic models for panoramic panoptic segmentation. Our findings indicate that models using well established baseline models such as the Seamless Scene Segmentation model~\cite{porzi2019seamless} which uses Mask R-CNN~\cite{he2017mask} do outperform more advanced transformer~\cite{vaswani2017attention} based models or fully convolutional panoptic models.

On a glance, we deliver the following contributions:
\begin{enumerate}
    \item We introduce the novel panoramic panoptic segmentation task, which aims to yield a $360^\circ$ holistic surrounding understanding, providing dense semantic- and instance information at the pixel level.
    \item We propose Panoramic Robust Feature (PRF), a contrastive-learning-driven panoramic panoptic segmentation framework with efficient architectures for mobile agents in road-driving scene parsing scenarios.
    \item We create WildPPS, the first panoramic panoptic segmentation dataset with images collected in cities located around the world, to foster robust surrounding perception. 
    \item The proposed method achieves state-of-the-art accuracy on both public PASS and our established WildPPS benchmarks. For a variety of CNN- and transformer-based panoptic segmentation models, our approach significantly improves their performances in PQ by $3.5{-}6.5\%$.
\end{enumerate}

\noindent\textbf{Difference from the Conference Paper:}
This paper is an extension of our conference work~\cite{jaus2021panoramic}.
Within this publication, we significantly increase the insights into the novel task introduced in our conference paper~\cite{jaus2021panoramic}, which received the Best Paper Award (Third Place) at the 2021 IEEE Intelligent Vehicles Symposium,
by adding the following contributions: 
\begin{enumerate}
    \item We double the number of annotated images forming the WildPPS dataset which results in more stable and reliable estimates for the performance of models.
    \item We add extensive experiments with multiple state-of-the-art panoptic segmentation models using various new adaptation techniques and extend the previously established efficiency setting to the new models.
    \item We add a benchmark of the proposed framework against a purely contrastive learning based training of the backbone.
    \item We take our proposed robust model to the test against multiple more powerful panoptic baseline models on WildPPS. 
    \item We provide novel insights into our established idea which combines supervised and contrastive training procedures suitable for a limited hardware setting.
    \item A more detailed description of the proposed framework and other enhanced parts such as related works and additional qualitative feature- and segmentation map analyses. 
\end{enumerate}

%% file: Tex_content/Related_work.tex
\subsection{Semantic, Instance, and Panoptic Segmentation}
Pixel-wise image, semantic- and instance-specific segmentation have advanced exponentially, driven by the architectural advances of deep Convolutional Neural Networks (CNNs)~\cite{he2017mask,he2016deep,long2015fully}.
The Fully Convolutional Network (FCN)~\cite{long2015fully} views semantic segmentation as an end-to-end pixel classification task, followed by encoder-decoder architectures~\cite{badrinarayanan2017segnet,chen2017deeplab,zhao2017pyramid} that significantly enhance segmentation performance by aggregating multi-scale contextual features.
Researchers further come up with promising approaches, by utilizing boundary cues~\cite{takikawa2019gated}, appending attention blocks~\cite{fu2019dual,yang2021capturing}, and leveraging vision transformers~\cite{zheng2021rethinking,xie2021segformer,zhang2022trans4trans} to improve FCN-based dense semantic understanding.
Additionally, there have been some attempts for semantic segmentation using supervised and unsupervised communication~\cite{liang2021cross_scene_foreground_segmentation} or leveraging knowledge distillation~\cite{liang2021nlkd,liu2022transformer_kd}.

Early instance segmentation models are built upon the ``detection followed by segmentation'' principle, with Mask R-CNN~\cite{he2017mask} being a well-known architecture, forming the basis of many contemporary box-based networks~\cite{liu2018path,bolya2019yolact}.
Recent box-free methods like SOLO~\cite{wang2020solo} and CondInst~\cite{tian2020conditional} devise FCN-like solutions, which offer simpler architecture with comparable segmentation performance to box-based models.
Moreover, additional methods tackle instance-specific image segmentation via clustering~\cite{arnab2017pixelwise,bai2017deep}, class-agnostic mask generation~\cite{pedro2015learning,dai2016instance}, and contour-based techniques~\cite{peng2020deep,xie2020polarmask}.

The lately introduced panoptic segmentation task~\cite{kirillov2019panoptic} unifies semantic- and instance segmentation in a single system, facilitating to recognize both things and stuff in urban driving scenes, which are of important relevance for autonomous vehicles.
A multitude of works~\cite{porzi2019seamless,xiong2019upsnet,li2020unifying} implement panoptic segmentation with a universal framework, showing the significance of the new task to driving scene parsing.
Among these approaches, some develop based on state-of-the-art instance segmentation methods like Panoptic FPN~\cite{kirillov2019pfpn} extending Mask R-CNN with a semantic branch for stuff segmentation~\cite{liu2019end,li2019attention}.
Another cluster of approaches enhances semantic segmentation architectures like Panoptic-DeepLab~\cite{cheng2020panoptic} extending DeepLab~\cite{chen2017deeplab}.
Further, there are works that capture the relations among semantic categories and instances for providing richer contexts to enhance visual understanding in panoptic segmentation~\cite{borse2022panoptic_relational}.

More recently, there are single-path architectures without using separate branches, which are realized via conditional convolution filters~\cite{hwang2022single,tian2022instance_panoptic}, learnable kernels~\cite{li2021panoptic_fcn,zhang2021knet}, and unified panoptic embeddings~\cite{kerola2021hierarchical}.
Transformer models have also been introduced into panoptic segmentation due to their capability to model long-range dependencies~\cite{vaswani2017attention,carion2020end,yu2022cmt}.
Inspired by DETR~\cite{carion2020end}, MaX-DeepLab~\cite{wang2020max} and MaskFormer~\cite{cheng2021maskformer} view unified image segmentation from a mask classification perspective.
MaX-DeepLab builds atop~\cite{wang2020axial} by extending the axial-attention backbone with a dual-path framework combining CNNs and transformers in the head network.
Building upon MaskFormer, Mask2Former~\cite{cheng2021masked} devises masked attention to extract localized features by constraining cross-attention with predicted mask regions.
Panoptic SegFormer~\cite{li2021panoptic_segformer} leverages an efficient deep supervision mask decoder and a query decoupling strategy to delve deep into panoptic segmentation.
In this work, considering that fast responses are critical for autonomous driving, we build on CNN- and transformer-based panoptic segmentation architectures~\cite{porzi2019seamless,li2021panoptic_fcn,cheng2021masked}. We make use of our efficient system and address panoptic segmentation in $360^\circ$ panoramic imagery to pursue a unified and complete surrounding understanding.

\subsection{Panoramic Scene Segmentation}
Modern scene segmenters are mostly designed to work with pinhole images on mainstream datasets such as Cityscapes~\cite{cordts2016cityscapes} and Mapillary Vistas~\cite{neuhold2017mapillary}.
To enlarge the Field of View (FoV), many surrounding understanding platforms are based on fisheye images or multiple cameras~\cite{deng2019restricted,yogamani2019woodscape,eising2021near_field_perception,kumar2021omnidet,Liao2021KITTI360}.
However, this either comes with severe distortions in particular around the fisheye image borders or leads to being cumbersome with a lot of multi-camera calibration work.
Motivated by the prospect of attaining $360^\circ$ perception with a single panoramic camera, recent works~\cite{zhang2019orientation,sekkat2020omniscape,orhan2021semantic_outdoor_panoramic} build directly on this modality, but they rely on synthetic collections that are far less diverse than pinhole databases~\cite{cordts2016cityscapes,neuhold2017mapillary}. A high variety of images is however critical for yielding robust segmentation models for real-world perception.

In contrast, the Panoramic Annular Semantic Segmentation (PASS) framework~\cite{yang2019pass} reuses knowledge in pinhole data to produce robust models suitable for $360^\circ$ images, and it is further augmented by DS-PASS~\cite{yang2020ds} via a detail-sensitive design with lateral attention connections.
In~\cite{yang2021capturing}, context-aware omni-supervised models are taken to the wild, fulfilling panoramic semantic segmentation in a single pass with enhanced generalizability.
P2PDA~\cite{zhang2021transfer} explicitly tackles panoramic segmentation from a domain adaptation perspective by transferring from the label-rich pinhole domain to the label-scarce panoramic domain.
In~\cite{hu2022distortion}, a distortion convolutional module is developed to correct the panoramic image distortion according to the image-forming principle.
Trans4PASS~\cite{zhang2022bending} architects a distortion-aware transformer with deformable token mixers for adapting to panoramic images.
However, existing $360^\circ$ perception systems only render semantic- or instance-specific segmentation, whereas a recent video panoptic segmentation work~\cite{mei2022waymo} based on the Waymo open dataset only offers a coverage of $220^\circ$. 
To achieve a unified and holistic scene understanding, this work advocates and first addresses panoramic panoptic segmentation, extending panoptic segmentation models with dense contrastive learning regimens that intertwine pixel-level consistency propagation for robust segmentation across pinhole- and panoramic imagery.

\subsection{Unsupervised Dense Contrastive Learning}
Currently, the most appealing approaches for learning representations without labels are unsupervised contrastive learning~\cite{he2020momentum,wu2018unsupervised,chen2020simple} tasks. Contrastive learning methods learn visual representations in a discriminative way by contrasting similar, positive pairs against dissimilar, negative pairs, which is promising to yield generalized features for robust predictions in previously unseen domains. The training pairs are often generated from augmented views of image samples, and thereby the previous methods are mostly designed for image classification tasks, which does not ensure more accurate pixel-wise segmentation~\cite{he2020momentum}.
Different from previous works, we aim to develop a contrastive learning regimen for pixel-level tasks.
Taking the wide FoV of omnidirectional data into consideration, in this paper, we put forward a learning framework for fine-grained panoptic segmentation operating on panoramic images towards a holistic understanding.

A few latest contrastive training methods~\cite{wang2020dense,xie2020propagate}, concurrent to our work, also address dense prediction tasks.
DenseCL~\cite{wang2020dense} implements self-supervision by optimizing a
pairwise contrastive (dis)similarity loss between two views of input images, whereas pixel-level pretext tasks are introduced for learning dense feature representations in~\cite{xie2020propagate}.
FisheyePixPro~\cite{cheke2022fisheyepixpro} attempts to pretrain a contrastive learning based model directly on fisheye images.
Cross-image pixel contrast has been leveraged for semantic segmentation by looking beyond single images~\cite{wang2021exploring,hu2021region,zhang2021looking} and enforcing pixel embeddings belonging to the same semantic class to be more similar than embeddings from different classes.
Some methods also explore pixel-to-region contrast~\cite{wang2021exploring,hu2021region,xiao2021region_similarity,liu2021bootstrapping} as a complementary to the pixel-to-pixel contrast strategy.
More recently, MaskCo~\cite{zhao2021self_contrastive_mask_prediction} introduces contrastive mask prediction for visual representation learning.
ORL~\cite{xie2021unsupervised_object_level} leverages image-level self-supervised pretraining as the prior to discover object-level semantic correspondence.
DSC~\cite{li2021dense_semantic_contrast} models semantic category decision boundaries at a dense level with multi-granularity contrastive learning.
In this work, we step further to open $360^\circ$ panoramic scenes and explore generalization effects from dense contrastive learning in a cost-minimizing way.

%% file: Tex_content/Proposed_Framework.tex
\label{Ch: Proposed Framework}
As the target of this work is to establish the most holistic surrounding understanding based on standard images, we are dealing with downstream tasks operating on a pixel level going beyond image level classification tasks. This observation is a key design principle we considered when mitigating the performance drop between the pinhole image source distribution and the panoramic target domain. Within this section, we show how to overcome the lack of annotated panoramic images by proposing the Panoramic Robust Feature (PRF) framework which generates robust features from pinhole images in a cost minimizing way, suitable for a mobile agent setting.

The proposed procedure consists of two steps in order to adapt panoptic target models.
The first step focuses on the backbone of the model which is pretrained in a pixel-level contrastive fashion, the second step is the standard supervised training of the panoptic model.

\begin{figure}
    \centering
    \includegraphics[trim=220 180 300 110, clip, width=0.4\textwidth]{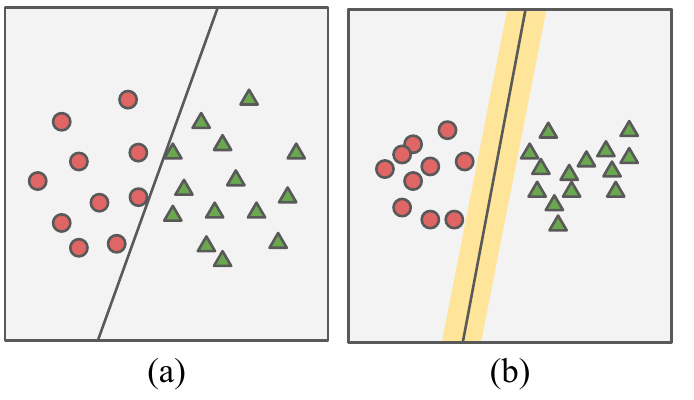}
    %\vskip-1ex
    \caption{Simplified representation of the feature embeddings obtained by a supervised training task via a linear classification (a) and a contrastive pretraining task (b). Supervised training features are only separated from each other to the extend which is necessary to maximize the performance of a linear classifier. It can be easily seen that the learned feature representation in (a) leads to the same classification result as the feature representation in (b) if the distribution of the data does not change by much. Via contrastive training, we do not only encourage similar features to be represented in a similar way but also dissimilar features to be represented as dissimilar as possible. As shown in (b), this leads to more robust features which can facilitate the downstream task in the face of distribution changes.}
    \label{fig:robust}
\vskip-3ex
\end{figure}

During the first step, the feature space, which was learned via the standard supervised ImageNet~\cite{deng2009imagenet} classification task, serves as a starting point.
Despite the proven ability of the commonly-used ImageNet weights for transfer learning tasks, we find that facing domain shift problems such as transferring from standard pinhole- to panoramic images causes severe performance drops. This behaviour may be caused by the fact that features learned via a final linear classification task are not encouraged to be well separated beyond what is necessary for a linear classifier, as shown in Fig.~\ref{fig:robust}.
Bearing in mind that intelligent vehicles often entail flexible model development and deployment, we aim to propose an approach which allows model modifications or retraining on a hardware setting which can run on a mobile agent.

Due to the limited availability of purely contrastive model backbones and the required access to at least $8$ GPUs~\cite{he2020momentum,xie2020propagate} to train them from scratch in order to achieve comparable results to the widely available ImageNet~\cite{deng2009imagenet} pretrained feature extractors, we unite the two approaches in our mixed training approach and take the best of both worlds. We use the widely available supervised ImageNet~\cite{deng2009imagenet} pretrained feature extractors and modify these weights using the contrastive task as described in Section~\ref{SS: Contrastive Pretraining Phase}. After adapting the backbone, a standard supervised training can be applied which will be introduced in Section~\ref{SS: Supervised Training Phase}. This makes the proposed Panoramic Robust Feature (PRF) framework a drop-in, model-agnostic procedure, which does not require changes in model architectures between pretraining and application-oriented fine-tuning. Furthermore, the required pretraining phase runs on a single GPU and finishes within very reasonable times, making our mixed training approach ideal for a limited hardware setting. 

\begin{figure*}
    \centering
    \includegraphics[trim=90 90 60 50, clip, width=0.9\textwidth]{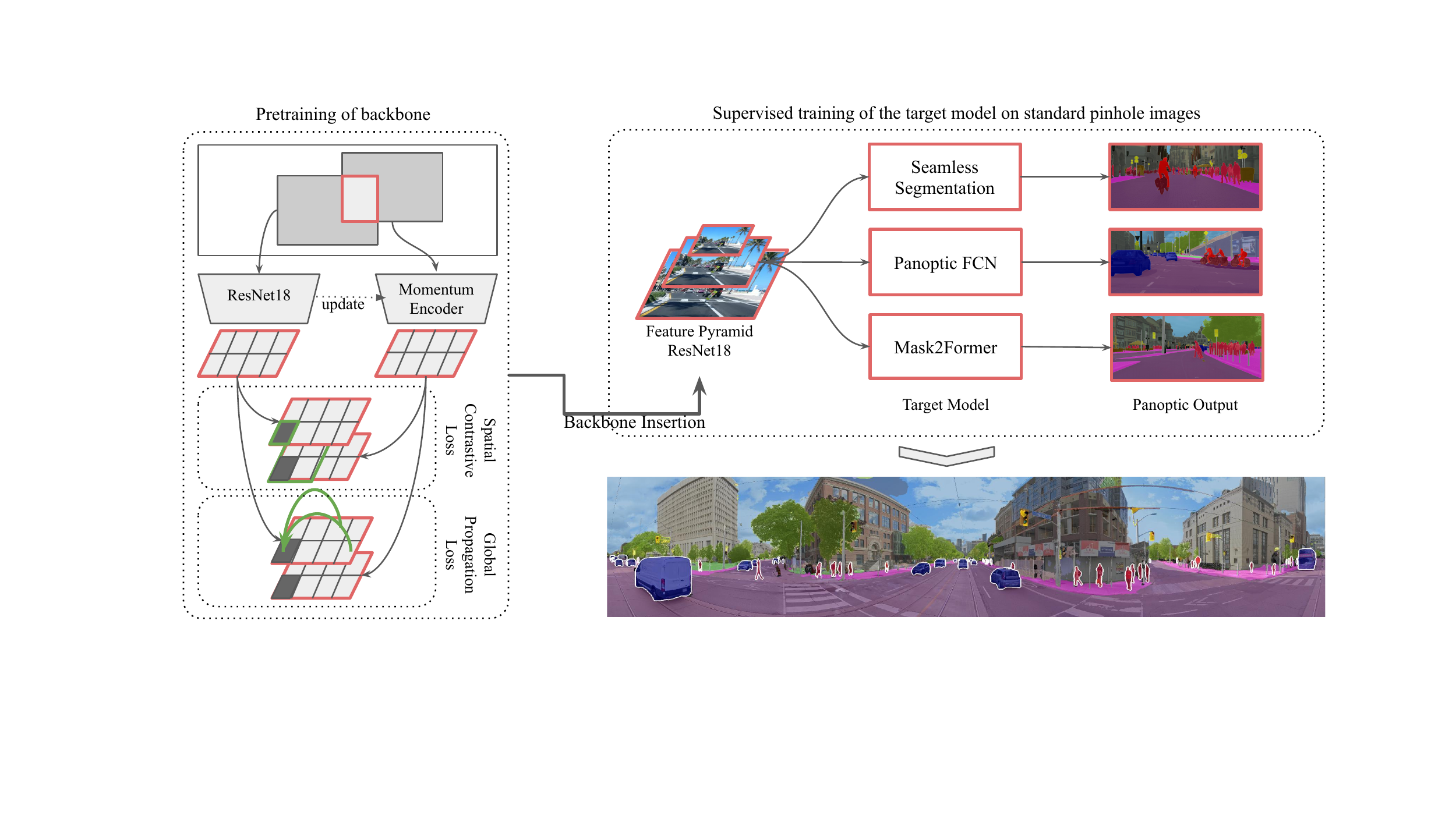}
    %\vskip-1ex
    \caption{The Panoramic Robust Feature (PRF) framework consists of two steps: A contrastive pretraining step as shown on the left side and the standard supervised training of the target model. During the pretraining, the features of the backbone are made robust by contrasting spatially close pixels against more distant pixels (green box), and globally-similar pixels are propagated onto each other. The robust backbone is inserted into the target model and the next step is to use the standard supervised training procedure on a pinhole image dataset such as Cityscapes~\cite{cordts2016cityscapes}. After training the robust model, it can produce high-quality panoptic annotations on our panoramic image dataset WildPPS. Within this work, we experiment with different target models in order to prove the effectiveness of the proposed approach and to identify models which are most suitable for panoramic panoptic segmentation.}
    \label{fig: framework overview}
    \vskip-3ex
\end{figure*}

\subsection{Contrastive Pretraining}
\label{SS: Contrastive Pretraining Phase}
The contrastive pretraining phase starts with the commonly-used supervised training weights $\theta_0^{enc}$ obtained by the $1000$ classes ImageNet~\cite{deng2009imagenet} classification task for a ResNet18 backbone~\cite{he2016deep}. 

From a given image, we crop two random views, augment them, and feed one of them through the encoder and the other through its momentum encoder~\cite{he2020momentum} initialized as $\theta_0=\theta_0^{enc}$. The augmentation techniques include standard procedures such as color jittering, flipping, gray-scale conversion, or solarization.
We use a standard ResNet18 encoder with a projection head on top. Following the work of~\cite{xie2020propagate}, the projection head is a simple network consisting of a $1{\times}1$ convolutional layer, projecting the features into a $2048$-dimensional latent space, a batch normalization layer, a ReLU nonlinearity layer, and a final $1{\times}1$ convolutional layer projecting the features to a target dimensionality of $256$.
We obtain two feature maps containing the features from the respective crops. The basic idea of the approach is that if the views overlap at a certain point, the feature maps of the corresponding pixels should be very similar. This idea builds upon the inherent translation equivariance of CNNs. 
After each step, we update the momentum encoder according to the following rule: 
\begin{equation}
    \theta_t = \theta_{t-1}\times\beta+(1-\beta)\times\theta_t^{enc}
    \label{Eq: Moment Encoder weight update}
\end{equation}
This update rule leads to an exponentially weighted moving average encoder of the regular encoder. Following~\cite{grill2020bootstrap}, we set $\beta{=}0.99$ which leads to both of the networks having similar weights, hence they should produce similar results for similar inputs which enables us to exploit two expected consistencies across the two feature maps.
We briefly explain them and their effect on panoramic panoptic image segmentation in this section and refer the interested reader to~\cite{xie2020propagate}.
Panoramic images are characterized by their wide FoV which requires the kernels of the backbone to analyze a plethora of different objects and pay attention to details as well as global contexts.

\noindent\textbf{Spatial Contrastive Loss:} 
%\newline 
This loss encourages the network to focus on the details of feature computations as it enforces the backbone to compute features that are similar for spatially close pixels and dissimilar for more distant pixels.
This is achieved by comparing two feature vectors $x_i$ and $x_i'$ computed from pixel $p_i$ by the regular encoder and the momentum encoder respectively.
If features originate from the same or spatially close pixels (indicated by the green frame in Fig.~\ref{fig: framework overview}) within the original image space, the features should be similar, despite their different locations within the two crops.
Consider the first cropped view $F$ and pixel $p_i$ located in $F$.
We define $\tilde{F}_i^{P}$ as the pixels located in the second view, which are spatially close to $p_i$ measured within the original image space.
The rest of the pixels in the second view are denoted by $\tilde{F}_i^N$ as their distance from $p_i$ exceeds a certain threshold.
$\tau$ is a normalization parameter set to $0.3$ as suggested by~\cite{xie2020propagate}.
The overall loss, as shown in Equation~\ref{Eq: Spatial contrastive loss}, is averaged across all pixels in each of the views, and the final loss is obtained as the average of the losses of the two views. 
\begin{equation}
L_{s}(p_i) = -\log\frac{\sum_{j \in \tilde{F}^P_i}e^{\frac{cos(x_i,x'_j)}{\tau}}}{\sum_{j \in \tilde{F}^P_i}e^{\frac{cos(x_i,x'_j)}{\tau}} + \sum_{k \in \tilde{F}^N_i}e^{\frac{cos(x_i,x'_k)}{\tau}}}
\label{Eq: Spatial contrastive loss}
\end{equation}

\noindent\textbf{Global Propagation Loss:}
%\newline 
This loss focuses on global consistency among similar pixels by propagating globally similar pixels onto the current pixel as shown by the green arrows in Fig.~\ref{fig: framework overview}.
This guarantees that the model assigns the same features to similar pixels beyond the FoV of the current kernel location,
making globally similar pixels semantically similar, which is in particular important for large-FoV panoramic images. An additional effect of this feature propagation is that it induces a smoothing effect, which prefers smooth outputs over fragmented ones as it averages out small differences. 

A smoothed feature vector $x_i^{smooth}$ can be calculated according to Equation~\ref{Eq: Smooth features} as a weighted sum consisting of all the projected features $g(x)$ in the entire view. The weights are determined by the similarity of the two respective feature vectors $x_i$ and $x_j$. 
\begin{equation}
    x_i^{smooth}=\sum_{j\in F} \max(\cos(x_i,x_j),0)^2*g(x_j)
    \label{Eq: Smooth features}
\end{equation}
Here, the transformation $g(x)$ can be computed via a $1{\times}1$ convolution keeping the number of channels constant.

Finally, the feature maps obtained by the encoder and the momentum encoder should produce consistent results for spatially close pixels $p_i$ and $p_j$ in the image space.
This behavior is encouraged by the global propagation loss shown in Equation~\ref{Eq: Glo Pro loss}.
\begin{equation}
    L_{GloPro}=-\cos(x_i^{smooth}, x_j') - \cos(x_j^{smooth},x_i')
    \label{Eq: Glo Pro loss}
\end{equation}

We find in our experiments that the additive combination of the Spatial Contrastive loss and the Global Propagation loss according to Equation~\ref{Eq: Final Loss} is very beneficial for our desired application and produces the expected smooth results. Unless stated otherwise, we set $\alpha{=}1$. More details about the experiments, ablation studies as well as qualitative results are shown in Sections~\ref{SS: Experimental Results} and~\ref{SS: Qualitative Analysis}.
\begin{equation}
    L_{Pretrain} = L_{s} + \alpha \times L_{GloPro}
    \label{Eq: Final Loss}
\end{equation}

\subsection{Supervised Training}
\label{SS: Supervised Training Phase}
One advantage of the proposed framework is its simple feasibility since it is a drop-in method that does not require any changes in the model architecture or the model training procedure. The only required changes may be due to the chosen mobile agent efficiency setting which will be briefly discussed in Section~\ref{SS: Experimental Setup}. The panoptic interpretation of the panoramic image can be obtained in a single forward pass and does not require any post-processing steps such as image stitching or fusion of different information sources.

The main design criteria of our approach are driven by the need for flexibility and efficiency. We want to establish a drop-in model-agnostic procedure for the pretraining as well as the supervised training phase. The proposed methods and models should work as efficiently as possible since energy is a valuable good in any mobile setting. We thus restrict the entire setup to a maximum number of two GPUs which we find a reasonable upper bound for a mobile agent.

The proposed efficient modifications of current state-of-the-art panoptic segmentation models follow our conference paper~\cite{jaus2021panoramic} and meet our established criteria, since we replace the standard ResNet50~\cite{he2016deep} with a ResNet18~\cite{he2016deep} backbone. This reduces the number of parameters in the backbone by $54\%$ from $25.6$ million to $11.7$ million but still remains a model-agnostic drop-in method. A more detailed efficiency comparison with the two different backbones plugged into the respective models as well as a comparison in inference speed are shown in Table~\ref{table_efficiency}. 

Our selection of panoptic target models is motivated by our goal to determine if certain types of models perform better than others in panoramic panoptic segmentation and to identify the most promising architectures to advance in this task. In order to provide a fair comparison, we make sure that all of the selected models are withing the same order of magnitude regarding the number of parameters, required Flops and processing time. We verify this in Table~\ref{table_efficiency} and Table~\ref{table_inferece_speed}.We differentiate between two-branch models combining the outputs of well-established instance and semantic segmentation models via a heuristic merging principle, single-branch models, and transformer-based models. The following models serve as representatives for the aforementioned categories:
\begin{itemize}
    \item \textbf{Seamless-Scene-Segmentation}~\cite{porzi2019seamless}\textbf{:} The seamless scene segmentation model combines two well-established baseline models: the Mask R-CNN~\cite{he2017mask} performing instance segmentation on the input and a DeepLabV3~\cite{chen2017rethinking} inspired second branch computes the semantic segmentation maps. Finally, a fusion step similar to~\cite{kirillov2019panoptic} combines the output of the two branches to the panoptic image. The model serves as an example of a two-branch model, which inherently treats things and stuff differently. 
    \item \textbf{Panoptic-FCN}~\cite{li2021panoptic_fcn}\textbf{:} Whereas previous models inherently treat things and stuff as different branches, the approach of this model is to treat them in a unified single-branch approach. It uses a kernel generator, which generates a dedicated convolutional kernel for each individual instance and each stuff region. Aside from the kernels, a high-resolution feature map is computed. The authors use the semantic branch of the Panoptic-FPN model~\cite{kirillov2019pfpn} as their feature encoder. The final panoptic results can be obtained via a convolution operation of the learned kernels with the encoded features. This eliminates the need for the final merging operation of the stuff and thing results as it was necessary in the two-branch models.
    \item \textbf{Mask2Former}~\cite{cheng2021masked}\textbf{:} Finally, we want to explore the capability of transformer-based models on panoramic panoptic image segmentation. We pick the Mask2Former model as a well-performing model with a general architecture which allows to perform semantic, instance, and panoptic segmentation without architectural changes. A transformer decoder in combination with an MLP computes the mask embeddings and class predictions for the masks based on the image features. The output of the model consists of a set of predicted masks, which are obtained by calculating the dot product between the mask embeddings and the encoded image features. Depending on the target task, the set of masks is post-processed to match the expected output format. 
\end{itemize}
The quality of the panoramic output is calculated according to the Panoptic Quality (PQ) measure~\cite{kirillov2019panoptic} as shown in Equation~\ref{Eq: Panoptic Quality}.
\begin{equation}
    PQ=\frac{\sum_{(p,g)\in TP}IoU(p,g)}{|TP|+\frac{1}{2}|FP|+\frac{1}{2}|FN|}
    \label{Eq: Panoptic Quality}
\end{equation}
Each of the efficient models generates panoramic panoptic segmentation in a single shot and is ready to be deployed in intelligent vehicle systems, delivering a complete and robust surrounding understanding. 

%% file: Tex_content/Experiments.tex
\subsection{Experiment Datasets}
\noindent\textbf{Cityscapes dataset.}
By doing pioneering work in the novel task of panoramic panoptic image segmentation, we are faced with two data-related challenges which result from the lack of panoramic images with panoptic annotations. As we cannot generate a dataset of sufficient size which can serve as a panoramic panoptic training dataset, we are forced to train on publicly available pinhole image datasets. Due to the wide acceptance in the field, we rely on Cityscapes~\cite{cordts2016cityscapes} for the supervised training procedure of the target models. Cityscapes is a dataset of street scene images captured under similar weather and lighting conditions in $50$ different locations in Germany and Switzerland. It consists of $2975$ training, $500$ validation, and $1525$ test images with fine panoptic annotations. 

\noindent\textbf{Mapillary Vistas dataset.}
Mapillary Vistas~\cite{neuhold2017mapillary} is a diverse street scene dataset which consists of $18k$ training, $2k$ validation, and $5k$ test images which are captured from all around the globe. The images have been collected under various weather and lighting conditions. In this work, we use the Mapillary Vistas dataset to benchmark our most promising models against previous state-of-the-art semantic segmentation models on the panoramic PASS dataset~\cite{yang2019pass}, as it uses the labeling scheme of Mapillary Vistas. 

\noindent\textbf{WildPPS: Wild Panoramic Panoptic Segmentation dataset.}
In order to evaluate the models on the panoramic panoptic segmentation task, we introduce WildPPS~\cite{jaus2021panoramic}, which is the first publicly available panoramic panoptic image dataset. In the latest version of WildPPS, we double the number of annotations, which results in $80$ panoramic images with fine panoptic annotations for the most important street scene classes. We provide annotations for the stuff classes \textit{street} and \textit{sidewalk} which are necessary to find traversable areas and the thing classes \textit{car} and \textit{person} that are essential in order to interact with other road users. WildPPS builds on top of WildPASS~\cite{yang2021context} and contains panoramic images of $40$ cities from all around the globe. It is a very diverse dataset containing street scenes not only from European or American cities but also from historically underrepresented regions such as Southeast Asia or Africa. The images are captured under various weather, lightning, and environmental conditions, which makes the dataset a challenging target domain. 
We adopt the Cityscapes annotation style in order to minimize the threshold to evaluate panoptic models on WildPPS. 
\begin{table*}[!t]
\caption{Overview of the main results of the proposed method. The respective target model is shown on the left whereas the respective robust adaptation technique is listed in the header.  
Bold numbers indicate the best performance for the respective target model.}
%\vskip-1ex
\label{table_mainresults}
\begin{center}

\begin{tabular}{|c|c|c|c|c|c|}
\hline
\multirow{2}*{\textbf{Panoptic Model}} & \multirow{2}*{\textbf{Baseline}} & \multicolumn{4}{c|}{\textbf{Panoramic Robust Feature (PRF) Approaches}} \\
\cline{3-6}
 & & \textbf{SGD} & \textbf{LARS} & \textbf{LARS Large} & \textbf{Pure} \\ \hline
\hline
Seamless Segmentation~\cite{porzi2019seamless} & 59.43\% & 63.24\% & 62.76\% & 61.54\% & \textbf{63.65\%}\\ \hline
Seamless Segmentation PQ Stuff   & 61.55\% & 68.10\% & 59.93\% & 55.84\% & \textbf{69.36\%}\\ \hline
Seamless Segmentation PQ Things  & 57.30\% & 58.38\% & 65.60\% & \textbf{67.25\%} & 57.93\%\\ \hline
\hline
Panoptic FCN~\cite{li2021panoptic_fcn}                & 49.29\% & 50.77\% &    44.82\%     &    41.26\%     & \textbf{52.93\%}\\ \hline
Panoptic FCN PQ Stuff       & 60.05\% & 62.39\% &    55.35\%     &    48.71\%     & \textbf{66.60\%}\\ \hline
Panoptic FCN PQ Things      & 38.53\% & 39.16\% &    34.28\%     &  33.80\%      & \textbf{39.26\%}\\ \hline
\hline
Mask2Former~\cite{cheng2021masked}        & 49.47\% & 54.93\% & \textbf{55.95\%} & 52.79\% & 51.91\%\\ \hline
Mask2Former Stuff  & 63.83\% & 69.60\% & \textbf{70.91\%} & 69.15\% & 69.69\%\\ \hline
Mask2Former Things & 35.11\% & 40.26\% & \textbf{40.99\%} & 36.43\% & 34.13\% \\ \hline
\end{tabular}
\end{center}
\vskip-3ex
\end{table*}

\subsection{Experimental Setup}
\label{SS: Experimental Setup}
\noindent\textbf{Baseline setups.}
As mentioned before, we use the Cityscapes dataset~\cite{cordts2016cityscapes} due to its wide acceptance within the community as the dataset of choice to perform the supervised training. All the experiments performed within this work were designed and executed with a mobile agent setting in mind. For the supervised training procedure, we restrict ourselves to the usage of two GPUs and otherwise follow the proposed training procedure of the authors of the respective models as closely as possible. The only noteworthy differences in the supervised training procedures result from adapting the original training procedures which typically rely on at least $8$ GPUs~\cite{porzi2019seamless,li2021panoptic_fcn,cheng2021masked} to a $2$ GPU setting, which is achieved by reducing the batch size and adapting the learning rate to the lower batch size. 
We experiment with the three different target models mentioned in Section~\ref{SS: Supervised Training Phase} and train them according to the previously discussed procedure. In order to assess the performance of the Panoramic Robust Feature (PRF) framework, a baseline model for each of the target models serves as a benchmark. The baseline model is simply the target model only trained in a supervised way on Cityscapes without a prior backbone adaption. The baseline models are evaluated on the panoramic WildPPS dataset and compared against multiple robust models.

\noindent\textbf{Pretraining settings.}
In order to generate robust models, we experiment with three different pretraining settings. All of the following pretraining experiments are performed on a single GPU and use the ImageNet pretrained weights as a starting point as described in Section~\ref{SS: Contrastive Pretraining Phase}. Reusing the knowledge in these weights reduces the adaptation time by two orders of magnitude compared to a purely contrastive training, which will be described later.

We achieve the best improvements of more than $6\%$ measured in PQ over the different non-adapted baseline models by pretraining on the same Cityscapes dataset on which we perform the supervised training. This eliminates the need for any other data than the one already available for the supervised training procedure. We train for $90$ epochs using a batch size of $100$ images. This procedure can be achieved on a single GPU by resizing the Cityscapes images to $256{\times}512$ pixels before feeding them into the framework.
\begin{itemize}
    \item Following~\cite{xie2020propagate}, we use a LARS optimizer~\cite{you2017large}, a base learning rate of $0.4$ and a cosine learning rate schedule with restarts every $30$ epochs. This serves as a regularization since the final model can be interpreted as an average across the restarts. We furthermore regularize the training by decaying the weights using a $1e^{-5}$ penalty. 
    \item The LARS optimizer was designed to work well with large batch sizes, hence we add a line of experiments in which the batch size is doubled to $200$. This setting will be referred to as LARS large. 
    \item In order to compare the LARS optimizer setting to a more standard one, we also make use of a SGD optimizer with momentum and a step learning rate scheduler in which we decrease the learning rate from $1e^{-3}$ by a factor of $10$ every $30$ epochs. The weight decay remains at $1e^{-5}$. 
\end{itemize}

\noindent\textbf{Purely contrastive counterpart.}
Finally, we want to benchmark our novel training approach which combines supervised and unsupervised training against a purely contrastive trained backbone. In order to train a ResNet18 backbone~\cite{he2016deep} from scratch, we perform our contrastive pretraining procedure starting with randomly initialized weights. In order to compare the contrastive backbone to the supervised backbones, we pretrain it on ImageNet~\cite{deng2009imagenet} for $100$ epochs with a batch size of $256$ images on $4$ GPUs using the LARS optimizer and a cosine learning rate schedule. Ideally, we would want to use $8$ GPUs, however the performance of the target models on the Cityscapes validation set is on par with that of their supervised counterpart, which makes us reasonably confident regarding the contrastive training. One exception is the Seamless Scene Segmentation model for which the pure backbone scores lower. We suppose that the supervised training procedure should be adapted to the contrastive backbone, but for a fair comparison, we keep the model. The total training time of this contrastive backbone was about $3.5$ days on $4$ GPUs compared to around $2h$ for the Panoramic Robust Feature (PRF) adaptation on a single GPU. 

\subsection{Experimental Results}
\label{SS: Experimental Results}

The main goal of this section is twofold: We want to verify the effectiveness of the proposed PRF framework. This is achieved by comparing the results of the robust versions of the target models to the non-adapted baseline models.
The second goal is to identify the most promising models to perform panoramic panoptic segmentation and answer the question if certain models are by design more suitable to excel at this novel and challenging task.  

\noindent\textbf{On the effectiveness of the PRF framework.}

The effectiveness of the PRF framework can be confirmed by comparing the WildPPS Panoptic Quality Scores of the Baseline models with those of the adapted models as shown in Table~\ref{table_mainresults}. 
Regarding the Seamless Scene Segmentation model~\cite{porzi2019seamless}, the effectiveness clearly can be confirmed as all of the different pretraining settings outperform the baseline model. The best performing adaptation is achieved by training with the purely trained backbone followed by the SGD setting. 

Regarding the Panoptic FCN model~\cite{li2021panoptic_fcn}, we find the same order of models outperforming the non-adapted target model. The LARS-adapted backbones do not work well and even worsen the performance. We conjecture, that the supervised training procedure of this model must be changed in order to profit from the LARS-adapted backbone features. The SGD-adapted backbone is, in contrast to the LARS-adapted backbone, still very similar to the original backbone as it is shown in Fig.~\ref{fig:Backbone comparison}, so the model training which has been optimized for the supervised baseline backbone still works quite well for the SGD backbone. The same line of arguments accounts for the purely trained backbone which produces more well-behaved embeddings without extreme feature isolation. We also investigate this behaviour in more detail within the qualitative analysis section. It seems that the FCN model tends to confuse zebra crossings with sidewalks, as shown in Fig.~\ref{fig:Qualitative results}.

Finally, the transformer-based Mask2Former model~\cite{cheng2021masked} can be improved reliably using the PRF framework. The best performing configuration is the LARS-based approach confirming that models that can capture the more heavily separated latent space modification benefit from it. 

Summarizing the presented results, we are confident regarding the effectiveness of the PRF framework. We suggest the SGD-based backbone as the default setting since we find it consistently outperforming the baseline and generating the second-best results in all settings using the training procedures proposed by the authors of the target models. In order to maximize the performance of a given target model, it can be beneficial to open up to the LARS or the Pure settings. We furthermore want to point out that in comparison to the purely contrastive backbone, we achieve comparable or better results using the orders of magnitude cheaper mixed training approaches, namely SGD, LARS, or LARS Large. This confirms the suitability of our framework for the desired mobile application setting.

\noindent\textbf{On the maximization of segmentation performance.}
The second posed question was on maximizing the performance of Panoramic Panoptic Segmentation. This question can be easily addressed, as the Seamless Scene Segmentation model~\cite{porzi2019seamless} outperforms any other model by a large margin, as it can be seen in Table~\ref{table_mainresults}, and provides the best score for PQ Things across all tested scenarios. The baseline model not only outperforms all the other baseline models but also the best-adapted versions of those models. We argue that this is largely due to the fact that the Seamless model uses two separate state-of-the-art models to compute the semantic segmentation map and the instance segmentation results, which are then combined into the final panoptic parsing result. These models have proven themselves in countless scenarios and have been updated and refined over the years, which makes them very robust by design. 

%%%%%%%%%%%%%%%%%%%%%%%%%%%%%StartLatelyIncluded%%%%%%%%%%%%%%%%%%%%%%%%%%
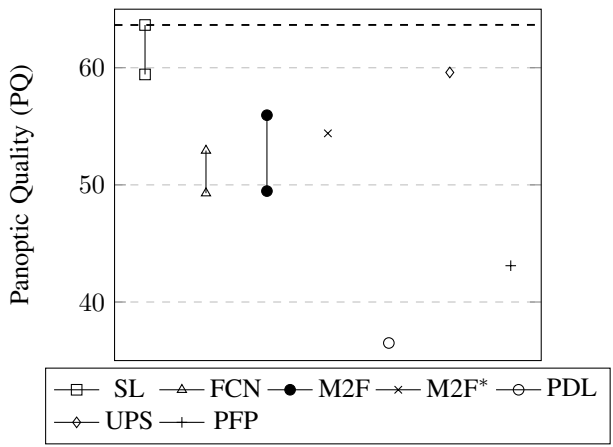
\begin{figure}[!t]
\centering
\begin{tikzpicture}
\begin{axis}[
    xlabel={},
    ylabel={Panoptic Quality (PQ)},
    xmin=0, xmax=7,
    ymin=35, ymax=65,
    xtick={63.65},
    ymajorgrids=true,
    grid style=dashed,
    width=0.4\textwidth,
    legend style={
    at={(0.5,-0.02)},
    anchor=north,
    legend columns=5},
]

\addplot[
    color=black,
    mark=square,
    ]
    coordinates {
    (0.5,63.65)(0.5,59.43)
    };

\addplot[
    color=black,
    mark=triangle,
    ]
    coordinates {
    (1.5,52.93)(1.5,49.29)
    };
    
\addplot[
    color=black,,
    mark=*,
    ]
    coordinates {
    (2.5,55.95)(2.5,49.47)
    };

\addplot[
    color=black,
    mark=x,
    ]
    coordinates {
    (3.5,54.4)
    };

\addplot[
    color=black,
    mark=o,
    ]
    coordinates {
    (4.5,36.5)
    };

\addplot[
    color=black,
    mark=diamond,
    ]
    coordinates {
    (5.5,59.6)
    };

\addplot[
    color=black,
    mark=+,
    ]
    coordinates {
    (6.5,43.10)
    };

\addplot[
    color=black,
    mark=None,
    dashed,
    line width=0.25mm
    ]
    coordinates {
    (0.0,63.65)(7,63.66)
    }
    ;

\legend{SL,FCN,M2F,M2F$^{*}$,PDL,UPS,PFP}
\end{axis}
\end{tikzpicture}
%\vskip-1ex
\caption{Comparison of our proposed efficient robust models with multiple panoptic baseline models. SL, FCN, M2F represent the Seamless-Scence-Segmentation~\cite{porzi2019seamless}, the Panoptic-FCN~\cite{li2021panoptic_fcn} and the Mask2Former~\cite{cheng2021masked} model. These are our efficient baseline models along with their robust adaptations. M2F$^{*}$, PDL, UPS and PFP represent the Mask2Former, Panoptic-DeepLab~\cite{cheng2020panoptic}, the UPSNet~\cite{xiong2019upsnet}, and the Panoptic-FPN~\cite{kirillov2019pfpn} model respectively. All of the latter models use a ResNet50 backbone~\cite{he2016deep} and are trained in a standard fashion.}
\label{fig: R2_comparison}
\vskip-3ex
\end{figure}

Regarding the remaining models, we cannot recommend the Panoptic FCN model~\cite{li2021panoptic_fcn} due to the low performance and the inability to cope with the modified LARS backbones. 

The Mask2Former model~\cite{cheng2021masked}, despite not performing well in its baseline scenario, can be improved to a solid $55.95\%$ PQ score and yields the best result for PQ Stuff among all the models, which might very well be attributed to the used attention mechanisms allowing global information exchange particularly relevant for the heavily distorted stuff classes. 

Nevertheless, our clear recommendation at this point is to use our robust Seamless Scene Segmentation model for domain shift problems.

\noindent\textbf{Comparison to panoptic baseline models.}
In order to prove the effectiveness of our robust efficient models against other panoptic models, we test the performance of standard panoptic models on WildPPS. We report the results in Fig.~\ref{fig: R2_comparison}. 
We trained the standard models as closely as possible to the settings proposed by the authors. For Panoptic-DeepLab, we replace the average pooling in the ASPP module with global average pooling in order to be able to use the model with changing image resolutions during inference. The performances of the baseline models on the Cityscapes validation dataset are comparable with each delivering about the same results as in the original papers. Their varying performances on WildPPS highlights once more the difficulty of the panoramic domain shift which not every model can cope with equally well. Not only is it visible that the proposed PRF framework yields consistent improvements over the non-adapted models, it furthermore can be seen that the proposed robust Seamless-Scene-Segmentation model outperforms a large variety of different models. This holds despite the fact, that the other baseline models were trained with a multiple of GPUs and use the more powerful ResNet50 backbone.

\begin{figure}[!t]
\centering
\begin{tikzpicture}
\begin{axis}[
    xlabel={$\alpha$},
    ylabel={Panoptic Quality (PQ)},
    xmin=0, xmax=4,
    ymin=60, ymax=63,
    xtick={0,1,2,3,4},
    ytick={61,62,64},
    ymajorgrids=true,
    grid style=dashed,
    width=0.3\textwidth
]

\addplot[
    color=black,
    mark=square,
    ]
    coordinates {
    (0.25,61.53)(0.5,62.24)(1,62.76)(2,62.06)(3,60.79)(4,61.06)
    };

\end{axis}
\end{tikzpicture}
%\vskip-1ex
\caption{Ablation study on the $\alpha$ hyper-parameter for the LARS setting. The x-axis indicates the weight between the Spatial Contrastive and the Global Propagation Loss as shown in Equation~\ref{Eq: Final Loss}.}
\label{fig: alpha}
\vskip-3ex
\end{figure}
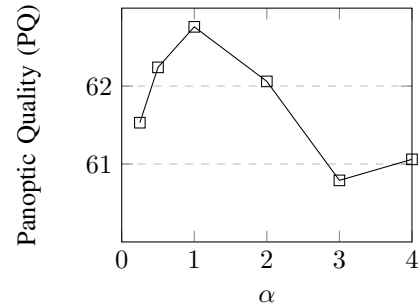

\noindent\textbf{Parameter study on $\alpha$.}
As discussed in Section~\ref{SS: Contrastive Pretraining Phase}, both the Spatial Contrastive loss and the Global Propagation loss play important roles in capturing both local as well as global image information. In order to determine the relative importance of the two losses, we perform an ablation study on $\alpha$ as introduced in Equation~\ref{Eq: Final Loss}. Following~\cite{jaus2021panoramic,xie2020propagate}, we use the LARS setting of the Seamless Scene Segmentation model~\cite{porzi2019seamless} to perform this study on the novel extended WildPPS dataset. The results are shown in Fig.~\ref{fig: alpha} and indicate that within reasonable bounds around $\alpha{=}1$, the results are quite similar. Deviating further from $1$ tends to decrease the performance. We obtain the best performance for exactly $\alpha{=}1$ which deviates from the results of the conference paper in which $\alpha{=}2$ performs best, however, due to the larger dataset used in the current work and the comparable results within $0.5{\leq}\alpha{\leq}2$, we believe that the performance within the stated interval is comparable and a more fine-grained hyper-parameter search could be restricted to the this range in order to maximize the performance. 

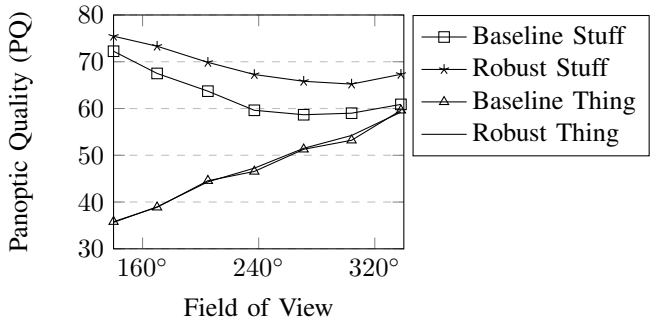
\begin{figure}[!t]
\centering
\begin{tikzpicture}
\begin{axis}[
    xlabel={Field of View},
    ylabel={Panoptic Quality (PQ)},
    xmin=140, xmax=340,
    ymin=30, ymax=80,
    xtick={160,240,320},
    xticklabel=$\pgfmathprintnumber{\tick}^\circ$,
    ytick={30,40,50,60,70,80},
    ymajorgrids=true,
    grid style=dashed,
    width=0.3\textwidth,
    legend pos=outer north east,
    legend cell align=left,
]

\addplot[color=black, mark=square] table[x=X, y=B_S]{Tex_content/pq.txt};
\addplot[color=black, mark=star] table[x=X, y=R_S]{Tex_content/pq.txt};
\addplot[color=black, mark=triangle] table[x=X, y=B_T]{Tex_content/pq.txt};
\addplot[color=black] table[x=X, y=R_T]{Tex_content/pq.txt};
\legend{Baseline Stuff, Robust Stuff, Baseline Thing, Robust Thing}

\end{axis}
\end{tikzpicture}
%\vskip-1ex
\caption{Detailed comparison of a standard Seamless Scene Segmentation model~\cite{porzi2019seamless} to its robust counterpart depending on the Field of View (FoV) on the WildPPS dataset.}
\label{fig: Detailed_robust_comparison}
\vskip-3ex
\end{figure}

\noindent\textbf{Generalization to different FoVs.}
Generally speaking, a difficulty when transferring from pinhole- to panoramic images is the increased Field of View (FoV) and the distortion resulting from mapping spherical image data to 2D planes. If the proposed robust modifications help to mitigate the targeted distribution shift from pinhole to panoramic images, we expect two behaviors: We should consistently outperform the baseline model across different FoVs and the performance gains should generally increase with increasing FoV. Both behaviors can be seen in Fig.~\ref{fig: Standard_robust_comparison}. As the increase in PQ with increasing FoVs is counter-intuitive, we explore this behavior in more detail by separating the Panotic Quality into the parts PQ Stuff and PQ Things. The results are shown in Fig.~\ref{fig: Detailed_robust_comparison}. As we crop starting from the center of the panorama, the Stuff areas tend to look similar to the Cityscapes dataset. By widening the FoV, the stuff areas get stretched across the entire panoramic FoV which is unusual, especially for the sidewalk class. The performance drop in PQ Stuff is mitigated by the robust backbone features. The counter-intuitive increase in PQ is caused by the increase in PQ Things which can easily be explained as cropping from the center in panoramic images tends to crop out small and distant objects, whereas the larger and easier to segment objects are on the side of the image. The difference in PQ Things of the two models is negligible, as cars and persons usually are present at different locations during training thus not leading to positional priors. On top, they are on average less affected by panoramic distortions due to their smaller surface area.

The presented results indicate that our model would also be potentially applicable and beneficial in other wide-FoV scenarios such as fisheye image segmentation and surround-camera perception, \textit{e.g.,} the $220^\circ$ sensing of the Waymo open dataset~\cite{mei2022waymo}, which are left for future exploration. 

\noindent\textbf{Generalization on Mapillary Vistas.}
Besides domain shifts from standard pinhole images to panoramic images, a related common problem is to have pinhole images obtained from a different distribution.
As this is not the main focus of this work, we just want to address this problem briefly within this section. We extend the results of our conference version~\cite{jaus2021panoramic} and add the novel LARS Large and Pure setting. Since Mapillary Vistas is a much more diverse dataset compared to Cityscapes, we map the learned Cityscapes labels to the corresponding Vistas label and ignore all the classes for which the model does not predict labels. The mapping is straightforward with two exceptions: We map the Cityscapes classes \textit{Rider} and \textit{Train} to the Mapillary classes \textit{Motorcyclist} and \textit{Other Vehicle} respectively. The results are shown in Table~\ref{table_mapillary} and indicate that not only the Seamless Scene Segmentation model is a very suitable model due to the inherent robustness for domain shift problems, but the results once more confirm the effectiveness of the PRF framework. We want to point out that the reported numbers in Table~\ref{table_mapillary} cannot be compared to the performance reports in the of the original Seamless-Scene-Segmentation paper. Our model was trained on Cityscapes and only performs inference on the classes available in Cityscapes which is only a fraction of the much more diverse Mapillary Vistas dataset.

\begin{table}[!t]
%\scriptsize 
\setlength{\tabcolsep}{2.0pt}
    \centering
    \caption{Comparison of results of the Seamless Scene Segmentation model~\cite{porzi2019seamless} on the Mapillary Vistas dataset~\cite{neuhold2017mapillary}.}
    %\vskip-1ex
    \begin{tabular}{|c|c|c|c|}
    \hline
        \textbf{Pretrain Setting} &  \textbf{PQ} & \textbf{PQ Stuff} & \textbf{PQ Things}\\
         \hline
         
         Baseline Model & 30.2\% & 38.6\% & 20.7\%  \\
         \hline
         \textbf{Pretrain Cityscapes SGD} & \textbf{31.9\%} & \textbf{41.7\%} & 20.7\% \\
         \hline
         Pretrain Cityscapes LARS & 31.3\% & 40.2\% & \textbf{21.2\% }\\
         \hline
         
         Pretrain Cityscapes LARS Large & 30.14\% & 38.31\% & 20.81\% \\
         \hline
         Pretrain Cityscapes Pure & 27.89\% & 34.26\% & 20.60\%\\
         \hline 
    \end{tabular}
    \label{table_mapillary}
    \vskip-3ex
\end{table}

\begin{table*}[!t]
%\scriptsize
\setlength{\abovecaptionskip}{0pt}
\setlength{\belowcaptionskip}{0pt}
\caption{Per-class accuracy analysis in Intersection over Union (IoU) and mean IoU (mIoU) on the public Panoramic Annular Semantic Segmentation (PASS) dataset~\cite{yang2019pass}.}
%\vskip-1ex
\label{table_pass}
\begin{center}
\begin{tabular}{|c|c|c|c|c|c|c|c||c|c|}
\hline
{\textbf{Network}}&{\textbf{Car}}&{\textbf{Road}}&{\textbf{Sidewalk}}&{\textbf{Crosswalk}}&{\textbf{Curb}}&{\textbf{Person}}&{\textbf{mIoU}}&\textbf{MACs}&\textbf{\#PARAMs}\\
\hline
\hline
{SegNet~\cite{badrinarayanan2017segnet}}&{57.5\%}&{52.6\%}&{17.9\%}&{11.3\%}&{11.6\%}&{3.5\%}&{25.7\%}&398.3G&28.4M\\
\hline
{PSPNet (ResNet50)~\cite{zhao2017pyramid}}&{76.2\%}&{67.9\%}&{34.7\%}&{19.7\%}&{27.3\%}&{22.6\%}&{41.4\%}&403.0G&53.3M\\
\hline
{DenseASPP (DenseNet121)~\cite{yang2018denseaspp}}&{65.8\%}&{62.9\%}&{30.5\%}&{8.7\%}&{23.0\%}&{8.7\%}&{33.3\%}&78.3G&8.3M\\
\hline
{DANet (ResNet50)~\cite{fu2019dual}}&{70.0\%}&{67.8\%}&{35.9\%}&{21.3\%}&{12.6\%}&{25.9\%}&{38.9\%}&114.1G&47.4M\\
\hline
\hline
{ENet~\cite{paszke2016enet}}&{59.4\%}&{59.6\%}&{27.1\%}&{16.3\%}&{15.4\%}&{8.2\%}&{31.0\%}&4.9G&0.4M\\
\hline
{CGNet~\cite{wu2020cgnet}}&{65.2\%}&{56.9\%}&{23.7\%}&{3.8\%}&{11.2\%}&{21.4\%}&{30.4\%}&7.0G&0.5M\\
\hline
{ERFNet~\cite{romera2017erfnet}}&{70.0\%}&{57.3\%}&{25.4\%}&{22.9\%}&{15.8\%}&{15.3\%}&{34.3\%}&30.3G&2.1M\\
\hline
{PSPNet (ResNet18)~\cite{zhao2017pyramid}}&{64.1\%}&{67.7\%}&{31.2\%}&{15.1\%}&{17.5\%}&{12.8\%}&{34.8\%}&235.0G&17.5M\\
\hline
{ERF-PSPNet~\cite{yang2019pass}}&{71.8\%}&{65.7\%}&{32.9\%}&{29.2\%}&{19.7\%}&{15.8\%}&{39.2\%}&26.6G&2.5M\\
\hline
{ERF-PSPNet (Omni-Supervised)~\cite{yang2021context}}&{81.4\%}&{\textbf{71.9\%}}&{\textbf{39.1\%}}&{24.6\%}&{26.4\%}&{44.1\%}&{47.9\%}&26.6G&2.5M\\
\hline
{SwiftNet~\cite{orsic2019defense}}&{67.5\%}&{70.0\%}&{30.0\%}&{21.4\%}&{21.9\%}&{13.7\%}&{37.4\%}&41.7G&11.8M\\
\hline
{SwaftNet~\cite{yang2020ds}}&{76.4\%}&{64.1\%}&{33.8\%}&{9.6\%}&{26.9\%}&{18.5\%}&{38.2\%}&41.8G&11.9M\\
\hline
\hline
{\textbf{Ours}}&{\textbf{84.7\%}}&{68.8\%}&{37.5\%}&{\textbf{50.2\%}}&{\textbf{27.4\%}}&{\textbf{61.4\%}}&{\textbf{55.0\%}}&115.2G&34.8M\\
\hline
\end{tabular}
\end{center}
\vskip-3ex
\end{table*}

\subsection{Comparison to Previous Models on PASS}
We compare our best model identified in Section~\ref{SS: Experimental Results} which is the Seamless-Scene-Segmentation model~\cite{porzi2019seamless} to previous works. As this work is the first to achieve panoptic segmentation on panoramic images, we can only compare it to semantic segmentation results on the publicly available Panoramic Annular Semantic Segmentation (PASS) benchmark~\cite{yang2019pass}. PASS includes $400$ unfolded panoramic annular images with pixel-level semantic annotations on $6$ navigation-relevant classes for evaluation with a resolution of $692{\times}2048$.

These compared networks, experimented by~\cite{yang2021context}, cover state-of-the-art accuracy-oriented segmentation models SegNet~\cite{badrinarayanan2017segnet}, PSPNet50~\cite{zhao2017pyramid}, DenseASPP~\cite{yang2018denseaspp}, and DANet~\cite{fu2019dual}.
Efficiency-oriented models are also included for a comprehensive comparative analysis, spanning ENet~\cite{paszke2016enet}, CGNet~\cite{wu2020cgnet}, ERFNet~\cite{romera2017erfnet}, PSPNet18~\cite{zhao2017pyramid}, ERF-PSPNet~\cite{yang2019pass},
SwifNet~\cite{orsic2019defense}, and SwaftNet~\cite{yang2020ds}.
They all view the wide-FoV panoramic image as a single input without separating it into multiple segments.
Under a fair comparison, we measure the Intersection over Union (IoU) of the different classes.
We also measure the complexity in MACs on an input resolution of $512{\times}1024$ and the number of parameters (\#Params) of our method by following the setup of~\cite{yang2020omnisupervised}.

Besides restricting ourselves to the semantic output of our model, we point out that our efficient modification is also among the most efficient models within the comparison.
We apply our proposed PRF framework to train a ResNet18 backbone followed by supervised training on Mapillary Vistas~\cite{neuhold2017mapillary}, as PASS is designed with the same labeling strategy as Vistas.
We follow the supervised training procedure described in Section~\ref{SS: Supervised Training Phase} for a total of $23$ epochs. Staring with an original learning rate of $1e^{-2}$, we reduce by a factor of $10$ after $95k$ and $125k$ iterations due to the observed loss saturation.
The results shown in Table~\ref{table_pass} illustrate that compared to the previous state-of-the-art segmentation networks, we surpass them by a large margin by using our proposed framework. In particular, the safety-critical class \emph{person} is well addressed by our model, outstripping the second-best model by $17.3\%$ in IoU. 
While our model is among the larger models, there are two important observations to make. Model size only seems to play an inferior role as our model does outperform larger models by a large margin and there are many small models which perform excellent. Secondly, model adaptation techniques are considerably more important than pure sizes. This can be seen as the second best model was adapted according to the ``Omni-Supervised'' scheme~\cite{yang2021context} and is rather small in size. The two adapted models do perform considerably better in heavily-distorted classes such as \emph{sidewalk} or \emph{curb}, proving the effectiveness of adaptation techniques for the panoramic domain shift. Our proposed model combines the advantages of a light-weight distortion-aware model with the performance of the latest object detection models such as Mask R-CNN~\cite{he2017mask} which explains the good performance on all of the classes.

\subsection{Efficiency Analysis}
\begin{table}
\centering
\caption{Efficiency comparison between the standard models and our proposed efficient modifications. As an efficiency benchmark, we use the number of parameters (in million parameters) and the number of GFLOPs on the training dataset Cityscapes~\cite{cordts2016cityscapes} and the evaluation dataset WildPPS. Images from both datasets are not resized and are fed at resolutions of $1024{\times}2048$ and $400{\times}2048$ for Cityscapes and WildPPS respectively.}
\label{table_efficiency}
\begin{tabular}{|c|c|c|c|} 
\hline
\textbf{Target Model}        & \textbf{Standard} & \textbf{Efficient} & \textbf{Savings}  \\ 
\hline
\hline
\multicolumn{4}{|c|}{Parameter Comparison in Millions}\\
\hline
\hline
Seamless Segmentation~\cite{porzi2019seamless} & 51.5             & 34.8               & 32.0\%              \\
\hline
Panoptic FCN~\cite{li2021panoptic_fcn}          & 36.8              & 23.7               & 35.6\%            \\ 
\hline
Mask2Former~\cite{cheng2021masked}           & 44.0                & 30.9               & 13.1\%            \\
\hline
\hline
\multicolumn{4}{|c|}{FLOPs Comparison on Cityscapes in GFLOPs}\\
\hline
\hline
Seamless Segmentation~\cite{porzi2019seamless} & 594.8 & 485.9   & 22.4\%  \\
\hline
Panoptic FCN~\cite{li2021panoptic_fcn} & 574.7 & 476.2 & 20.7\%\\
\hline
Mask2Former~\cite{cheng2021masked} & 527.4 & 419.6 & 25.7\% \\
\hline
\hline 
\multicolumn{4}{|c|}{FLOPs Comparison on WildPPS in GFLOPs}\\
\hline
\hline
Seamless Segmentation~\cite{porzi2019seamless} & 282.3 & 239.5 & 17.9\% \\
\hline
Panoptic FCN~\cite{li2021panoptic_fcn} & 233.5 & 193.5 & 20.7\%\\
\hline
Mask2Former~\cite{cheng2021masked} & 215.1 & 171.3 &  25.6\% \\
\hline
\end{tabular}
\vskip-3ex
\end{table}

\noindent\textbf{Computation savings.}
Regarding the targeted mobile agent application, it is essential to work with hardware- and energy-efficient models. Our proposed efficient model modification approach reduces both, the number of parameters and the required FLOPs, which serve as a proxy for hardware- and energy efficiency respectively, while it still remains a model-agnostic drop-in method which does not require any architectural changes. This allows a straightforward adaption of arbitrary target models, such that even future models should be easily adaptable and deployable. In Table~\ref{table_efficiency}, we compare the standard models as proposed by the respective authors with our efficient modifications. Depending on the setting and the target model, we can reduce the number of parameters up to $36\%$ and decrease the required FLOPs to calculate the panoptic predictions up to $26\%$.

\begin{figure}
    \centering
    \begin{tikzpicture}
    \datavisualization [scientific axes={end labels},
   x axis={label={Image Multiplier}, unit length={25mm per 1 units}, ticks=few},
   y axis={label={Saved GFLOPs}},
   visualize as scatter]
    data {
       x, y
1.1 ,	38
1.2 ,	46
1.3 ,	55
1.4 ,	65
1.5 ,	75
1.6 ,	86
1.7 ,	97
1.8 ,	111
1.9 ,	123
2.0 ,	138
2.1 ,	151
2.2 ,	168
2.3 ,	184
2.4 ,	201
2.5 ,	218
2.6 ,	238
2.7 ,	256
2.8 ,	277
2.9 ,	297
3.0 ,   320
    };
\end{tikzpicture}
\caption{Difference in GFLOPs between the standard ResNet50 backbone~\cite{he2016deep} and the efficient ResNet18 backbone~\cite{he2016deep} with increasing image size. The x-axis denotes the scalar with which the height and the width of the original WildPPS~\cite{jaus2021panoramic} resolution are multiplied. On the very left, the difference is below $50$ GFLOPs at around $400{\times}2048$ pixels and increases to over $300$ GFLOPs for images three times the resolution of WildPPS.}
\label{fig: Flop savings}
\vskip-3ex
\end{figure}
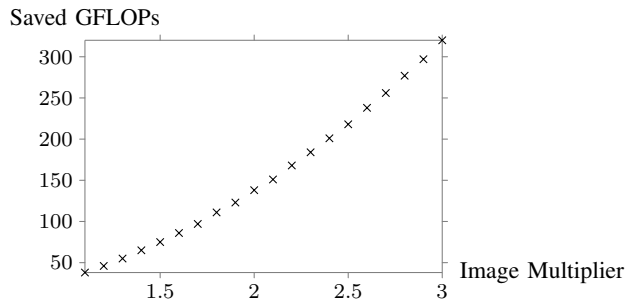

As the required FLOPs depend on the size of the input image which can only be expected to increase in the future, we want to point out that the difference in FLOPs between the standard models and our efficient modifications will increase in line with the number of pixels. This relationship is visualized in Fig.~\ref{fig: Flop savings}, where we show the difference in GFLOPs between a ResNet50~\cite{he2016deep} and a ResNet18~\cite{he2016deep} depending on the number of input pixels. We start from the original WildPPS resolution of $400{\times}2048$ pixels and increase the number of pixels threefold to $1200{\times}6144$ pixels while keeping the typical panoramic aspect ratio constant. This is achieved by multiplying the height and width with the same scalar. This scalar is shown on the x-axis of the graph. Despite the fact that there is more to the difference in FLOPs than the difference in the backbones of the models, this is a realistic proxy, indicating that our efficient modifications become even more important when feeding images of higher resolutions into the model.

\begin{figure*}%
    \centering
    \includegraphics[trim= 32 25 30 35, clip, width=0.95\textwidth]{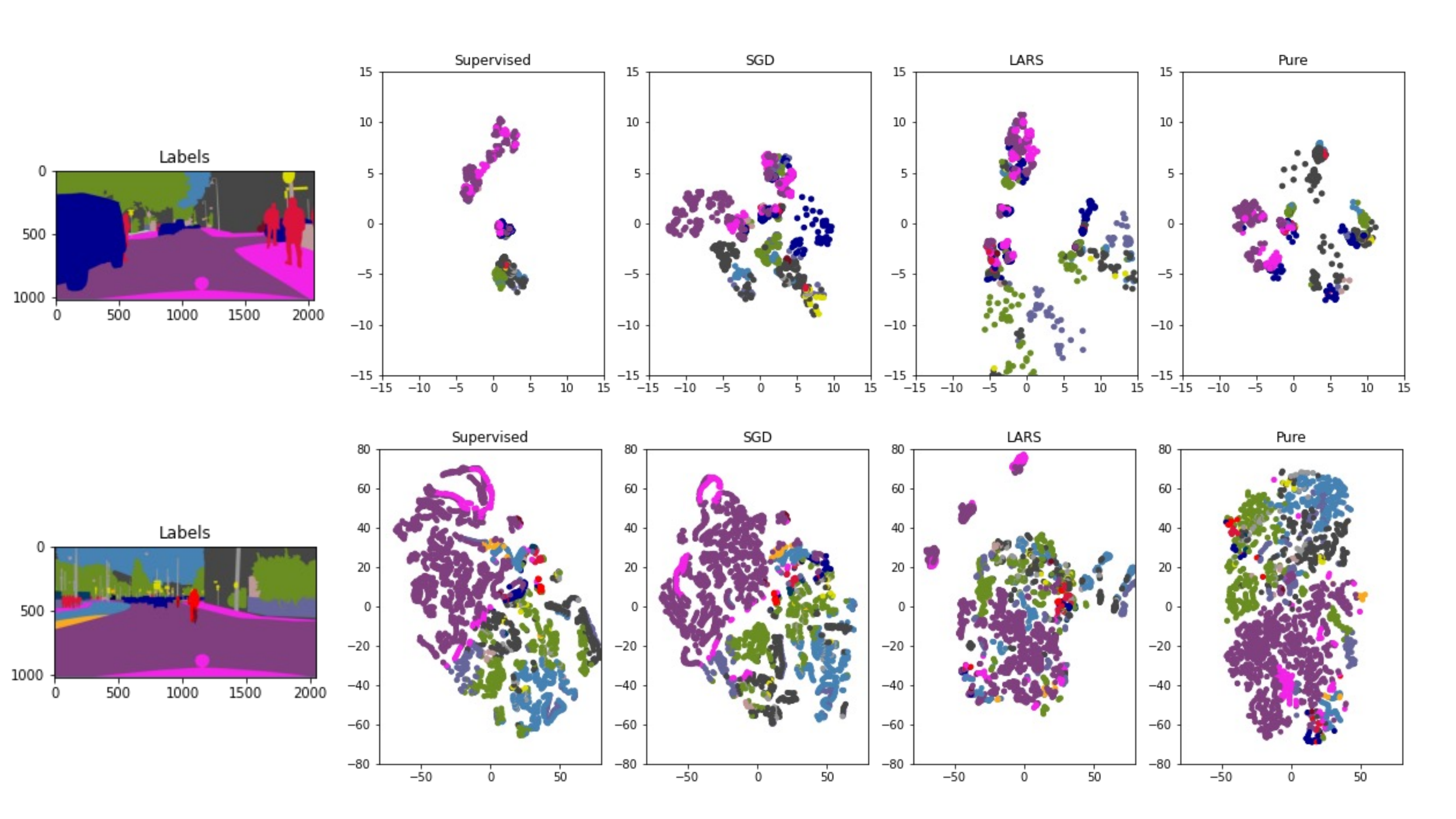}%
    \caption{Comparison of the latent space representations of the standard supervised backbone as shown in the left column and the different forms of backbone adaptions in the following columns. SGD, LARS, and Pure denote the mixed trainings according to the SGD and LARS setting as described in Section~\ref{SS: Experimental Setup} and the purely contrastive backbone respectively. In the first row we use the novel h-NNE visualization technique~\cite{sarfraz2022hierarchical}, whereas in the bottom row we rely on the established t-SNE dimensionality reduction~\cite{van2008visualizing}. For a better comparison among the different adaptation techniques, we set the axis to a reasonable interval. Best viewed on a screen and with zoom.}%
    \label{fig:Backbone comparison}%
    \vskip-3ex
\end{figure*}

\noindent\textbf{Inference speed.}
Besides hardware and energy efficiency, inference speed is a further necessary criterion in order to assess the suitability of an image parsing system for mobile agents. This section addresses this issue by measuring the necessary time to calculate a panoptic street scene prediction for a given single image at full resolution. The reported results, as shown in Table~\ref{table_inferece_speed}, assume the model has been built and is located on the GPU. The reported times include the necessary time to ship a given image to the GPU and the execution time to generate the panoptic result including possible necessary post-processing steps such as the heuristic merging procedure used by the Seamless Scene Segmentation model. The experiments have been conducted on a single GeForce RTX 2080Ti GPU and the numbers denote the average inference time of the entire validation dataset.
\begin{table}[h]
    \centering
    \caption{Inference speed of the proposed efficient target models measured as the average Frames Per Second (FPS) on the two relevant image datasets using a single GeForce RTX 2080Ti GPU.}
    \label{table_inferece_speed}
    \begin{tabular}{|c|c|c|}
    \hline
         Model & FPS on Cityscapes & FPS on WildPPS \\
         \hline
         \hline
         Seamless Segmentation~\cite{porzi2019seamless} & 4.4 & 8.0\\
         \hline
         Panoptic FCN~\cite{li2021panoptic_fcn} & 5.5 & 9.1\\
         \hline
         Mask2Former~\cite{cheng2021masked} & 4.5 & 9.0\\
         \hline
    \end{tabular}
    \vskip-3ex
\end{table}

Comparing the inference time between Cityscapes and WildPPS, the Cityscapes images are slower as they contain about $2.5$ times as many pixels.
Regarding the inference times on WildPPS, which reflects the desired scenario of a mobile agent operating on panoramic images, our efficient model modifications are able to process between $8$ to $9$ Frames Per Second (FPS). This is below the common $24$ FPS threshold, thus the models only offer near real-time performance. There are several possibilities to address this issue. Besides reducing the input resolution which would increase the inference speed but likely decrease the performance, a further step towards real-time processing would be in line with our experimental setup in which we used two GPUs. Using the training setup for inference could double the FPS by asynchronous image processing. In a two GPU scenario, we are confident that with minor inference modifications or by using slightly more powerful GPUs, a $24$ FPS real-time performance is possible for all the proposed models.

\subsection{Qualitative Analysis}
\label{SS: Qualitative Analysis}

\noindent\textbf{Feature space analysis.}
As stated in Section~\ref{Ch: Proposed Framework}, an intuitive explanation regarding the functionality of the Panoramic Robust Feature (PRF) framework is the adaptation and the restructuring of the latent space obtained by the supervised ImageNet classification tasks. In this section, we visualize the introduced adaptations by comparing the backbones used in the baseline model to the robust backbones and the purely contrastive backbone. We use the output of the last pooling layer of the ResNet18 backbone~\cite{he2016deep}, which result in a $32{\times}64$ pixel resolution feature map with $512$ filters which we map to a $2$-dimensional visualization space using the novel h-NNE~\cite{sarfraz2022hierarchical} and the well-established t-SNE~\cite{van2008visualizing} reduction technique.

\begin{figure*}%
    \centering
    \includegraphics[trim=0 0 385 30, clip, width=0.95\textwidth]{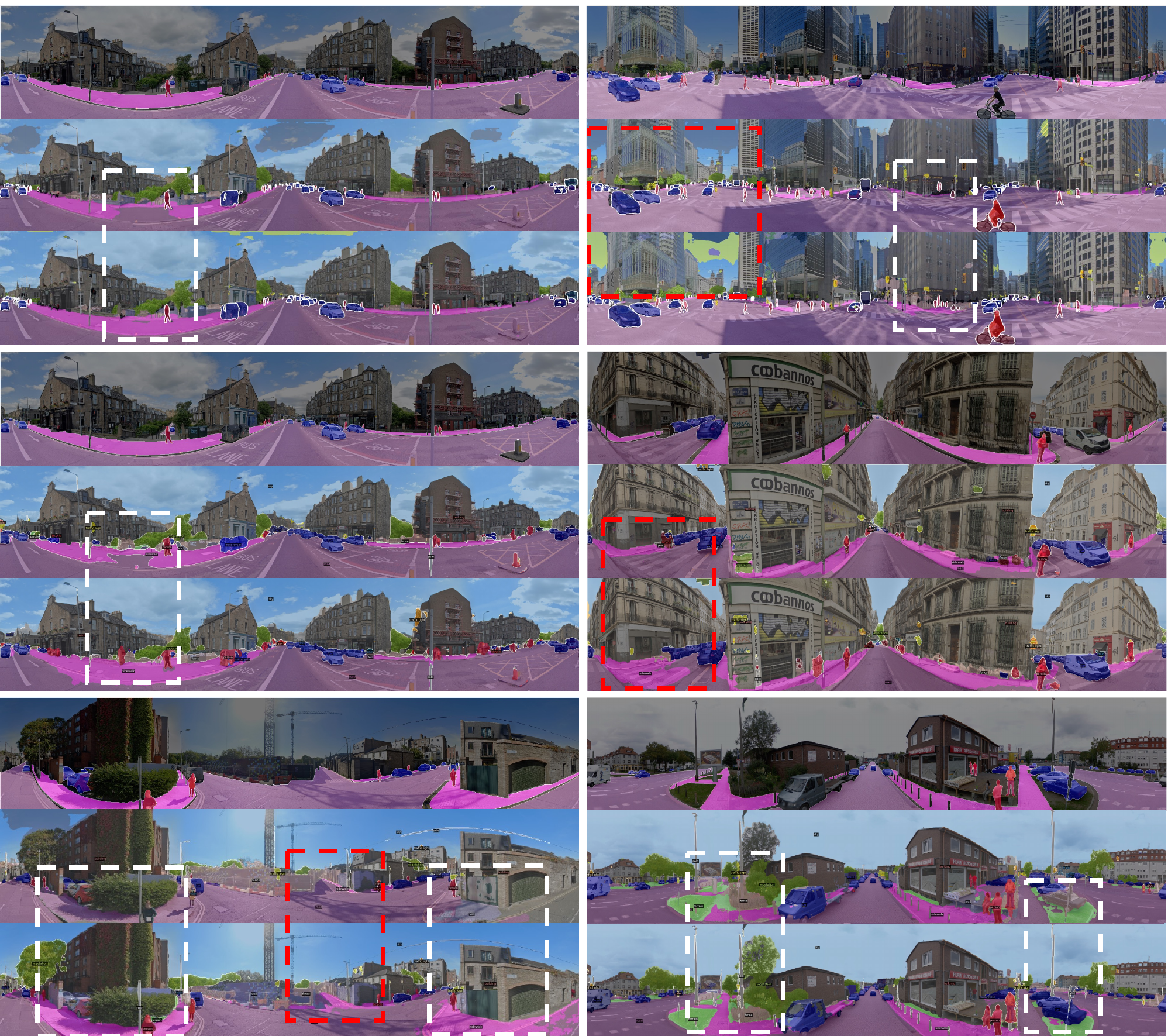}%
    \caption{Qualitative comparison of the baseline models with their robust counterparts. In each of the image triplets, the first image shows the manually annotated ground truth, followed by the baseline model, and finally the robust model. The first, second, and third row show examples of the Seamless Scene Segmentation~\cite{porzi2019seamless}, the Panoptic FCN~\cite{li2021panoptic_fcn}, and the Mask2Former~\cite{cheng2021masked} model, respectively. Triplets in the first column show results adapted according to the SGD setting, whereas images in the second column show results of the target models adapted via the LARS setting. White boxes show improvements of the robust models. In red we highlight undesired behaviors. Best viewed on a screen and with zoom.}%
    \label{fig:Qualitative results}%
    \vskip-3ex
\end{figure*}

The results are shown in Fig.~\ref{fig:Backbone comparison} and support the stated intuition. Regarding the h-NNE visualization plots which are shown in the first row, the different adaptation techniques pull apart the features of different labels while features representing the same labels still remain close to each other. Comparing the LARS setting to the SGD setting, we observe that the label separation is performed more aggressively by the LARS optimizer. The same effect can also be seen in the t-SNE plot in the second row. We hypothesize that this larger deformation of the latent space has the potential to improve the results even further compared to the SGD setting, if the model can cope with these changes. For the chosen target models, the Mask2Former model seems to be able to capitalize on this, whereas the Panoptic FCN model is not. 

The t-SNE visualization technique as shown in the second row provides more insights into the structure within the different labels. Comparing the supervised to the SGD setting, the latent space structure is very similar and this setting only slightly changes the feature encoder.

It is very plausible that the supervised training procedures of the used target models are heavily optimized to maximize the performance on the commonly used supervised backbone features. We observe that the SGD setting being the closest to the original latent space is a reliable choice that outperforms the baseline consistently as shown in Table~\ref{table_mainresults}. The LARS setting on the other hand has the potential to outperform the SGD backbone if the model is able to cope with this setting or if the training procedure is fine-tuned to this backbone.

\noindent\textbf{Segmentation map analysis.}
Finally, we compare the segmentation maps of the baseline models to those of the robust modifications, in order to qualitatively assess the effect of the Panoramic Robust Feature framework. The segmentation maps are shown in Fig.~\ref{fig:Qualitative results}.
It can be easily seen that PRF enables both CNN- and transformer-based panoptic models to produce complete and more reliable panoramic surrounding understandings.
The most present and apparent effect is that the robust features help in mitigating positional priors, which are most common for the \emph{sidewalk} class that is typically only present at very similar locations during the supervised training phase. The robust features help to overcome these positional priors and help to improve the sidewalk segmentation. Sometimes this behavior however does go beyond the desired effect, marking zebra crossings or unusual roads as sidewalks. From an application point of view, this behavior, although not desirable, poses less dangers to the environment compared to not being able to capture a sidewalk at all. 
Some of these behaviors remain unpunished at the moment such as the reported behavior in the second column of the first row, leaving room for further improvements.

%% file: Tex_content/Conclusion.tex
%Conclusions
In this work, we have introduced the novel task of panoramic panoptic image segmentation which is the most holistic scene understanding task based on standard camera input. Only by combining a panoptic image-level understanding with a panoramic field of view, a mobile agent operating in a real-world scenario such as traffic scenes is able to make informed decisions. Semantic Segmentation maps are well suited to identify traversable areas such as roads, whereas instance masks can capture the different traffic participants, hence panoptic image segmentation perfectly combines the advantages of both approaches. 
We examine different architectures of Panoptic Segmentation models in order to identify the most promising types for the proposed task. As these models are complex and require a substantial amount of labeled data to be trained, we train them on publicly available large-scale image datasets such as Cityscapes. We address the difficult distribution change from pinhole- to panoramic images which is mitigated by the proposed Panoramic Robust Feature framework that leverages a pixel-level contrastive pretext task. The framework allows the training of robust models in a cost-minimizing way, suitable for the targeted mobile agent application. We confirm the effectiveness of the proposed approach by evaluating the target models on the WildPPS dataset, the first panoramic panoptic dataset which will be published in order to foster progress in panoramic panoptic scene parsing. 
The Seamless Scene Segmentation model proves to be a reliable choice for this new task achieving state-of-the-art performance on WildPPS and the PASS dataset.
Finally, we confirm the efficiency of both the proposed panoramic models and the adaptation framework in the face of hardware and energy scarcity to meet the mobile agent setting.

%Future work 
Since the proposed method does not only generalize well to panoramic images but also generates improvements when train and test dataset differ as shown in Table~\ref{table_mapillary}, we believe that the proposed PRF framework has the capabilities to produce robust models in particular suitable for domain shift problems. Due to the desired real world applications of mobile agents, we think our approach has the potential to perform well in other mobile agent vision related sensing methods such as LiDAR or RADAR and their corresponding tasks such as segmentation of point clouds. 
As the pinhole to panoramic domain shift is a known problem for the aforementioned sensing modalities, we believe that robust PRF models could help in overcoming these domain shift as well.

%Wrap it up
We hope that this work sparks interest in the community regarding Panoramic Panoptic Segmentation. In the future, we want to examine different distribution shifts not only from pinhole- to panoramic images but also from artificially generated panoramic images to real-world panoramic images. Furthermore, we are curious about the effects of the framework when including more annotated classes as some of the undesired behaviors are not yet punished due to the lack of labels for the classes. This may open up additional research questions as different classes are affected differently by the panoramic distortions and may require special treatment.